\pdfoutput=1
\documentclass[11pt]{article}

\usepackage{EMNLP2023}

\usepackage{times}
\usepackage{latexsym}

\usepackage[T1]{fontenc}
\usepackage[utf8]{inputenc}

\usepackage{microtype}

\usepackage{inconsolata}

\usepackage{mathtools}
\usepackage{lipsum}% http://ctan.org/pkg/lipsum
\usepackage{caption, subcaption}
\usepackage{url}
\usepackage{amsmath, amsfonts, amsthm, bm}
\usepackage{soul}
\usepackage{tabularx, multirow}
\usepackage{dsfont}
\usepackage{booktabs}
\usepackage{xcolor}
\usepackage{chngcntr}
\usepackage[ruled,vlined]{algorithm2e}
\usepackage{algpseudocode}
\usepackage{graphicx}% http://ctan.org/pkg/graphicx

\usepackage{xspace}

\newcolumntype{X}{>{\centering\arraybackslash}m{0.2\linewidth}}
\newcolumntype{Y}{>{\raggedright\arraybackslash}m{1.0\linewidth}}

\newcommand{\proposed}{\textsc{RTSum}\xspace}

\newenvironment{talign*}
 {\csname align*\endcsname}
 {\endalign}

\newcommand{\bart}{BART\xspace}
\newcommand{\tfive}{T5\xspace}

\newcommand{\sen}[1]{{s}_{#1}\xspace}
\newcommand{\rel}[1]{{r}_{#1}\xspace}
\newcommand{\phr}[1]{{p}_{#1}\xspace}

\newcommand{\senedge}{{E}^{\text{s}}\xspace}
\newcommand{\reledge}{{E}^{\text{r}}\xspace}
\newcommand{\phredge}{{E}^{\text{p}}\xspace}

\newcommand{\senscore}{{S}^{\text{s}}\xspace}
\newcommand{\relscore}{{S}^{\text{r}}\xspace}
\newcommand{\phrscore}{{S}^{\text{p}}\xspace}

\newcommand{\cnndm}{CNN/DM\xspace}

\newcommand{\senset}{\mathcal{S}}
\newcommand{\relset}{\mathcal{R}}
\newcommand{\phrset}{\mathcal{P}}

\newcommand{\smallsection}[1]{{\vspace{0.03in} \noindent \bf {#1.\hspace{5pt}}}}

\title{\proposed: Relation Triple-based Interpretable Summarization with~Multi-level Salience Visualization}

\author{
  Seonglae Cho, Myungha Jang, Jinyoung Yeo, Dongha Lee\thanks{\ \ Corresponding author}\\
  Yonsei University, Republic of Korea \\
  \texttt{{\{sungle3737,jinyeo,donalee\}@yonsei.ac.kr}, {myunghajang@gmail.com}}  \\
}

\begin{document}
\maketitle
\begin{abstract}

In this paper, we present \proposed, an unsupervised summarization framework that utilizes relation triples as the basic unit for summarization. 
Given an input document, \proposed first selects salient relation triples via multi-level salience scoring and then generates a concise summary from the selected relation triples by using a text-to-text language model. 
On the basis of \proposed, we also develop a web demo for an interpretable summarizing tool, providing fine-grained interpretations with the output summary.
With support for customization options, our tool visualizes the salience for textual units at three distinct levels: sentences, relation triples, and phrases. 
The code\footnote{\texttt{https://github.com/seonglae/RTSum}} and video\footnote{\texttt{https://youtu.be/sFRO0xfqvVM}} are publicly available.

%Most unsupervised approaches to extractive summarization aim to select a subset of sentences that cover key information of a source document, without the help of human annotated text-summary pairs for model training. Despite their effectiveness, the selection of a whole sentence cannot exclude unnecessary or peripheral information from the sentence, and this limits the abstractiveness and flexibility of the output summary. In this work, we focus
%on a relation triple, which describes a smaller
%piece of information than a sentence, and study
%how to leverage them as the basic unit for %summarization. The proposed RTSum framework
%(1) selects the salient relation triples by fully
%utilizing heterogeneous textual units identified
%in a source document and then (2) generates
%plausible sentences from the selected relation
%triples by employing a text-to-text languagehttps://www.overleaf.com/project/64b237d8a068cfd379f933da
%model optimized in a self-supervised way. Our
%empirical evaluation demonstrate that RTSum
%achieves not only higher ROUGE scores than
%state-of-the-art extractive methods but also better factual consistency against the source document compared to recent abstractive methods
\end{abstract}

\section{Introduction}

Text summarization has emerged as a critical tool in the era of information overload, enabling users to quickly understand the essence of long text. 
Among various summarization techniques, abstractive summarization has gained significant attention due to its ability to generate fluent and concise summaries that capture the main ideas of the source text \cite{nallapati-etal-2016-abstractive, see-etal-2017-get, tan-etal-2017-abstractive, cohan-etal-2018-discourse, xu-etal-2020-understanding-neural, koh-etal-2022-far}. 
Nevertheless, despite their advantages in flexibility and reduced redundancy compared to extractive methods, abstractive methods inherently lack interpretability.
% The generation of new sentences can sometimes lead to semantic drift, where the summarized information deviates from the original meaning. 
That is, the absence of a direct link to the source text can make it difficult to trace back the source of information, which makes the summary lack interpretability. 

Interpretability in summarization is important to provide users a way to cross-check that the generated summary is factually consistent, and to provide more context to dive into if one wants to know more about the summarized content. To generate an interpretable summary, extractive summarization techniques can offer advantages.
As they directly extract sentences from the text, the sentences themselves serve as the source of information \cite{xu-etal-2020-unsupervised, padmakumar2021unsupervised}. 
However, a significant drawback of many extractive methods lies in their sentence-level operation, which limits their ability to extract fine-grained key information ~\cite{zheng-lapata-2019-sentence, lu2021unsupervised}.
% However, the majority of extractive techniques operate at the sentence level, selecting whole sentences from the source text to include in the summary. 
% These extractive methods, based on sentence selection, are limited in that they are not able to extract fine-grained key information. 
In many cases, a single sentence describes multiple diverse pieces of information that should be treated as distinct facts for summarization. 
% Extracting the entire sentence can lead to summaries that include unnecessary or redundant information, thereby reducing the efficiency and readability of the summary.
By selecting entire sentences, these methods may include unnecessary or redundant information in the summary, reducing both its efficiency and readability.

To enhance the interpretability of the summarization process by incorporating fine-grained key information, our focus lies on leveraging \textit{relation triples} as the basic unit for summarization. 
A relation triple in the form of (subject, predicate, object) concisely describes a single piece of information corresponding to its relation (i.e., predicate) between two entities (i.e., subject and object), and it can be effectively identified from a source document by using open information extraction (OpenIE) systems~\cite{angeli-etal-2015-leveraging, mausam2016open}.

Using relation triples, our main idea embodies \textit{selection-and-sentencification}, which achieves a combination of extractive and abstractive summarization methods.
Specifically, we first select only a few relation triples according to their importance -- \textit{salience} -- within the document for summarization, and then reassemble the selected relation triples into the final output summary.
This two-step approach enhances the interpretability of the summarization by providing clear explanations for the salience scores of relation triples and their contributions to the final summary.
This clarity allows users to understand the crucial elements driving the summarization process effectively.
% empowers us to better 
% Each sentence is decomposed into multiple relation triples, which are finer-grained information units than sentences, and only salient ones that mostly cover key contents of the source document are selected.
% Then, the selected relation triples are reassembled into plausible sentences, thereby obtaining a final output summary.
% the redundancy of source sentences is reduced, and simultaneously, the output summary is forced to be factually consistent with the source document.

Formally, we present an unsupervised \underline{R}elation \underline{T}riple-based \underline{Sum}marization framework, named \proposed.
For relation triple selection, \proposed identifies heterogeneous textual information units with various granularity, which are
(1) sentences, (2) relation triples, and (3) phrases, to utilize their own salience all together.
Under the principle that more salient textual units are much more relevant to other units semantically and lexically, it models the multi-level salience from the three distinct textual units.
Then, it selects the $K$ most salient relation triples based on the multi-level salience scores.
For relation triple sentencification, 
\proposed employs a neural text-to-text architecture as a \textit{relation combiner} to transform the relation triples into the summary sentences.
The relation combiner is effectively optimized in a self-supervised manner by using source sentences (sampled from training documents) and their relation triples (extracted from the sampled sentences) as targets and inputs, respectively, while not requiring any reference summaries of its training documents.

% Our tool, operating at a phrase level, extracts the most relevant pieces of information from the source text, resulting in more concise and informative summaries. 
% By focusing on relation-based triples, our tool can capture the key relationships and facts within the text, providing a more detailed and nuanced summary than traditional sentence-level extractive methods.

Building upon the \proposed framework, we develop an online demo to showcase an interpretable text summarization tool.
Given an input document, our tool generates a concise summary, while simultaneously offering fine-grained interpretations by visually depicting the multi-level salience of textual units within the source document.
For clarity in visualization, the tool highlights text spans (i.e., textual units) based on their salience score, with numerical ranks provided as annotations. 
Furthermore, our tool offers customization options, allowing users to personalize the visualization according to their preferences and specific purposes.

Our multi-level salience visualization empowers users to easily identify the textual units that mostly influence the final summary;
it also provides valuable insights into the salient semantic structure of the document at a glance, enhancing users' overall understanding of the summarization process.

\label{sec:intro}

\section{Preliminary}

\begin{figure*}[thbp]
    \centering
    \includegraphics[width=\linewidth]{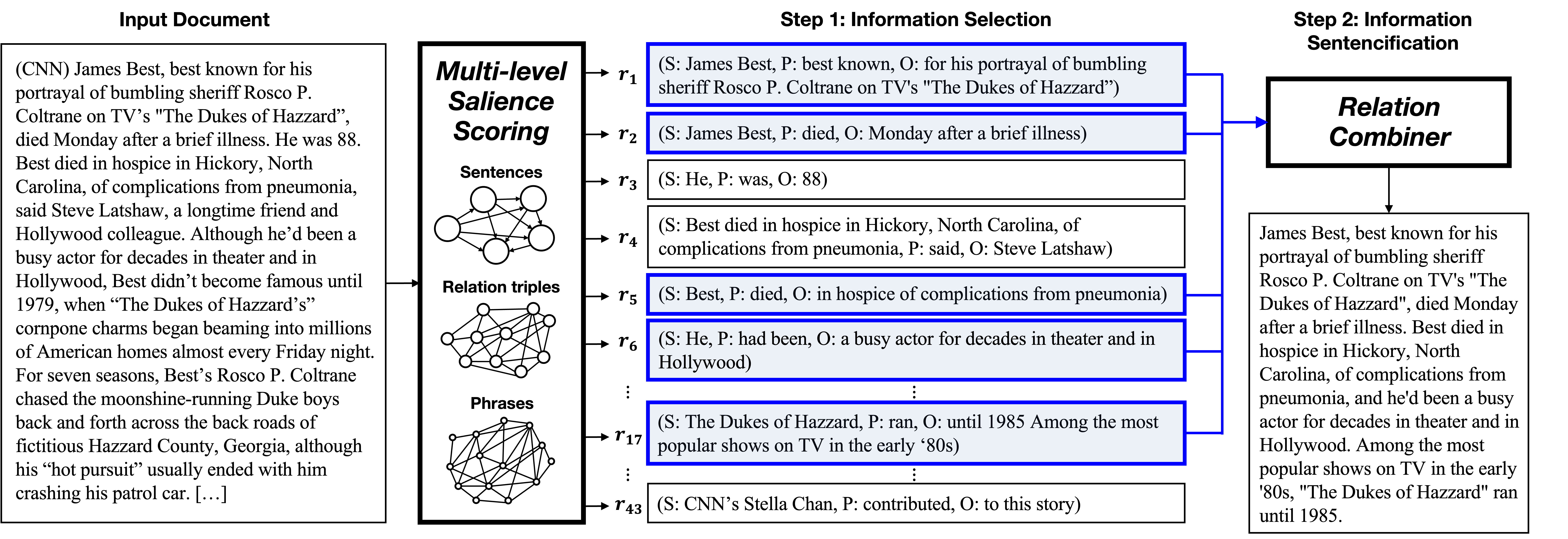}
    \caption{The overall process of the \proposed framework. \proposed selects salient relation triples and then generates the plausible sentences from the selected relation triples.}
    \label{fig:rtsum}
\end{figure*}

\smallsection{Textual Information Units}
We utilize three different types of textual information units with various granularity:
sentences, relation triples,\footnote{For brevity, we use the terms ``relation triples'' and ``relations'' interchangeably, in the rest of this paper.} and phrases.
All sentences and relations are extracted from a source document by using open information extraction (OpenIE) systems.
Among several implementations, we employ OpenIE 5\footnote{\texttt{https://github.com/dair-iitd/OpenIE-standalone}}~\cite{mausam2016open} released by UW and IIT Delhi.
Similarly, all noun and verb phrases in the document are identified based on POS labels tagged by the Spacy library.
Table~\ref{tbl:informunits} shows an example of the three textual information units in a single sentence.

\begin{table}[htbp]
    \centering
    \resizebox{0.99\linewidth}{!}{%
    \begin{tabular}{XY}
    \toprule
    Sentence &  Hugh Laurie joins the cast and Julia Louis-Dreyfus is now the president of the United States on HBO's hit comedy. \\ \midrule
    \multirow{3}{*}{Relation} & (S: Hugh Laurie, P: joins, O: the cast) \\
    & (S: Julia Louis-Dreyfus, P: is, O: now the president of the United States on HBO's hit comedy) \\ \midrule
    Phrase & Hugh Laurie, joins, cast, Julia Louis-Dreyfus, is, president, United States, HBO's, hit comedy\\
    \bottomrule
    \end{tabular}
    }
    \caption{An example of textual information units.}
\label{tbl:informunits}
\end{table}

% this openie system also has an advantage of not being canonicalized, which results in redundant relations -> which can prevents from some information loss during the extraction process.

\smallsection{Relation Triples}
Each relation triple, denoted by $\rel{}=(\text{sub},\text{pred},\text{obj})$, represents a relation (i.e., predicate) between two text spans (i.e., subject and object), and it corresponds to a single piece of information in terms of the relation.
The three components are described in natural language, and this allows us to treat them as the sequence of tokens in the vocabulary, similar to sentences and phrases.
Thus, we consider the concatenated text of its subject, predicate, and object as the textual description of a relation triple, i.e., $\text{desc}(\rel{})=[\text{sub}\Vert\text{pred}\Vert\text{obj}]$.

\smallsection{Problem Definition}
Given a source document $D$ and its information units, including sentences $\senset=\{\sen{1}, \ldots, \sen{N_s}\}$, relation triples $\relset=\{\rel{1}, \ldots, \rel{N_r}\}$, and phrases $\phrset=\{\phr{1}, \ldots, \phr{N_p}\}$, the goal of our relation triple-based summarization task is (1) to select the relation triples based on their salience within the document, and (2) to generate a concise summary from the selected salient relation triples.
In this paper, we mainly focus on the unsupervised setting where the annotated text-summary pairs are not available for training a summarization model, since such reference summaries are usually noisy, expensive to acquire, and hard to scale.

\label{sec:prelim}

\section{\proposed: Relation Triple-based Summarization Framework}

Our summarization framework, named \proposed,
%mainly utilize relation triples extracted from a source document as the basic information unit to obtain a concise summary of the document.
%The summarization process 
consists of the two steps: (1) \textit{information selection} for identifying the salient relation triples based on multi-level salience from various textual information units (Section~\ref{subsec:stepone}), and (2) \textit{information sentencification} for combining the selected relation triples into plausible sentences with the help of a neural text generator (Section~\ref{subsec:steptwo}).
Figure~\ref{fig:rtsum} illustrates the overall process of our \proposed framework.

\subsection{Information Selection}
\label{subsec:stepone}
For the selection of relation triples that would be included in the summary, 
\proposed models the multi-level salience for each relation triple by leveraging heterogeneous textual information units.
Specifically, it figures out how significant a relation triple is within the document from the perspective of (1) the sentence that the relation triple is extracted from, (2) the relation triple itself, and (3) the phrases that the relation triple contains.
% \proposed models the salience of a relation at three different levels, each of which figures out how significant the relation triple is within the document from the perspective of sentences, relations, and phrases, respectively.
% \proposed models the salience of each sentence, relation triple, and phrase based on semantic relation among their relation
% To this end, it first identifies the information units of different granularity from an input source document, and then constructs a multi-level text graph that represents the semantic relevance among three levels of textual semantic units: 
% That is, for each input document, a three-level text graph is constructed as illustrated in Figure~\ref{fig:textgraph}.
In the end, \proposed selects the most $K$ salient relation triples based on their multi-level salience scores.\footnote{The selection (or ranking) strategy based on multi-level salience can be implemented in various ways (Section~\ref{subsubsec:relselection}).}
% That is, it progressively checks the salience for relation triples, inferred from its relevant coarse-grained to fine-grained units. 

\subsubsection{Sentence-level Salience Score}
\label{subsubsec:sensal}
The sentence-level salience considers the significance of the sentence that each relation triple is extracted from.
Following previous studies~\cite{zheng-lapata-2019-sentence, lu2021unsupervised},
% \citet{zheng-lapata-2019-sentence, lu2021unsupervised} empirically demonstrated that modeling the sentence order (or position) along with the semantic similarity between the sentences is effective to identify key sentences.
we infer the sentence-level salience by utilizing sentence order (i.e., a preceding sentence is more likely to contain salient information) and semantic similarity (i.e., the sentence that is more semantically relevant to other sentences is likely to contain salient information).
Thus, we construct a sentence-level text graph $\mathcal{G}^\text{s}=(\senset,\senedge)$ with a directed edge from a former sentence node $\sen{i}$ to a latter sentence node $\sen{j}$, and the edge weight $\senedge_{ij}$ is the semantic similarity between the two sentences.
\begin{equation}
\label{eq:senedge}
    \senedge_{ij} = 
    \begin{cases}
    \text{sim}(\sen{i}, \sen{j}) & \text{if $\sen{i}$ precedes $\sen{j}$}\\
    0 & \text{otherwise}.
    \end{cases}
\end{equation}
$\text{sim}(\sen{i}, \sen{j})$ is defined by the cosine similarity between two sentence representations from a sentence encoder, specifically fine-tuned for the semantic textual similarity (STS) task~\cite{gao-etal-2021-simcse}.

From the sentence-level text graph, the sentence-level salience is defined by the degree-based centrality~\cite{zheng-lapata-2019-sentence}.
In other words, this centrality is equivalent to the sum of semantic similarities with all of its subsequent sentences.
\begin{equation}
\label{eq:senscore}
    \senscore(\sen{i}) = \sideset{}{_{s_j\in\senset}}\sum \senedge_{ij}.
\end{equation}

\subsubsection{Relation-level Salience Score}
\label{subsubsec:relsal}
The relation-level salience focuses on the meaning of each relation itself in that the semantic similarity among the relation descriptions implies the salience;
that is, a relation description that is more relevant to other relation descriptions is more likely to contain salient information. 
In this sense, we build a relation-level text graph $\mathcal{G}^\text{r}=(\relset,\reledge)$, whose nodes represent the relation triple and the undirected edge has the weight of the semantic similarity between relation descriptions.
Similar to Equation~\eqref{eq:senedge}, the cosine similarity between relation representations is calculated, and the salience score is also modeled as the degree-based centrality.
\begin{equation}
\label{eq:relscore}
\begin{split}
    \reledge_{ij} &= 
    \text{sim}(\text{desc}(\rel{i}), \text{desc}(\rel{j})), \\
    \relscore(\rel{i}) &= \sideset{}{_{\rel{j}\in\relset}}\sum \reledge_{ij}.
\end{split}
\end{equation}
Note that the sequential order of relations is not clearly presented unlike the sentences, because multiple relations are extracted from the same sentence.

\subsubsection{Phrase-level Salience Score}
\label{subsubsec:phrsal}
The phrase-level salience measures the salience of phrases included in each relation triple, and it captures the phrase frequency and co-occurrence in the document.
As presented in previous work on keyphrase extraction~\cite{mihalcea-tarau-2004-textrank, bougouin-etal-2013-topicrank}, we build a phrase-level text graph $\mathcal{G}^\text{p}=(\phrset, \phredge)$ whose nodes are the noun and verb phrases extracted from the source document based on POS tags (e.g., \texttt{Noun}, \texttt{Proper Noun}, and \texttt{Verb}).
The undirected edges model the weight as how many times two phrases locally co-occur (i.e., within a sliding window) in the sentences $\senset$, and then run the TextRank~\cite{mihalcea-tarau-2004-textrank} on the graph to compute salience of phrase nodes.
\begin{equation}
\label{eq:phrscore}
\begin{split}
    \phredge_{ij} &= 
    \text{co-occur}(\phr{i}, \phr{j}; \senset), \\
    \phrscore(\phr{i}) &= (1-d) + d \cdot 
    \sum_{\phr{j}\in\phrset}  \frac{\phredge_{ji}}{\sum_{\phr{k}\in\phrset} \phredge_{jk}} \phrscore(\phr{j}),
\end{split}
\raisetag{55pt}
\end{equation}
where $d\in[0, 1]$ is the damping factor that indicates the transition probability from one node to another random node.
%, and the edge weight is defined by the co-occurrence frequency of two phrases in each sentence $\sen{}\in\senset$.
Starting from initial values of $\phrscore$ usually set to 1.0 for all the nodes, the final salience of each phrase is obtained through iterative computation of Equation~\eqref{eq:phrscore} until convergence.

\subsubsection{Salient Relation Triple Selection}
\label{subsubsec:relselection}
The remaining challenge here is to select relation triples by integrating multi-level salience scores.
%: $\senscore$, $\relscore$, and $\phrscore$.
To this end, \proposed first identifies the textual information units relevant to each relation triple $\rel{i}$, including its source sentence $\sen{j}$ and its phrases $\phr{k} (\in\phrset_{\rel{i}})$, and then transforms their salience scores for the relation triple by
% \begin{itemize}
%     \item $\senscore(\rel{i}) := \senscore(\sen{j})$ where $\sen{j}$ is the source sentence of $\rel{i}$
%     \item $\relscore{\rel{i}} := \relscore{\rel{i}}$
%     \item $\phrscore(\rel{i}) := \sum_{\phr{k}\in\rel{i}}\phrscore(\phr{k})$.
% \end{itemize}
$\senscore(\rel{i}):=\senscore(\sen{j})$, $\relscore(\rel{i}):=\relscore(\rel{i})$, and $\phrscore(\rel{i}):=1/|\phrset_{\rel{i}}|\cdot \sum_{\phr{k}\in\phrset_{\rel{i}}} \phrscore(\phr{k})$.

The most straightforward strategy to select a small number of salient relation triples is to calculate the final score of each relation triple based on \textit{weighted summation} of its three distinct scores and to select the top-$K$ relation triples:
\begin{equation}
    S(r_i) = \alpha\cdot S^\text{s}(r_i) + S^\text{r}(r_i) + \beta\cdot S^\text{p}(r_i).
\end{equation}

Another selection strategy is to adopt \textit{cascade filtering} that excludes less salient relation triples by using the sentence-level, relation-level, and phrase-level salience in a serial order (i.e., $S^\text{s}\rightarrow S^\text{r}\rightarrow S^\text{p}$).
The key principle of this filtering process is to keep only the relation triples extracted from the key sentences, and among them, to selectively collect the relation triples that are semantically relevant to the others, and finally, to exclude the ones that do not include many salient phrases.
% Note that this approach does not aggregate the three scores, which eventually reduces the efforts to tune the hyperparameters for weighted summation or weighted product.
% Compared to sentence extractive methods, it is able to further eliminate the redundant relational information from the selected sentences by leveraging relation-level and phrase-level salience scores. 

% \smallsection{Weighted Ranking}
% The final saliency score of each relation triple $\rel{i}$ is defined by
% \begin{equation}
% S^{\text{final}}(\rel{i}) = \alpha \cdot \nsenscore(\sen{j}) + \beta \cdot \nrelscore(\rel{i}) + \gamma \cdot \nphrscore(\phr{k}),
% \end{equation}
% where $\alpha$ and $\beta$ are the hyperparameters for weighting its belonging sentence and inclusive phrases, respectively.
% All the extracted relation triples in $\mathcal{R}$ are sorted by their final saliency score, and the $K$ most salient triples are selected to be included in the output summary.
% Note that $\alpha$ and $\beta$ are manually set, or easily tuned by using a validation set.
% To normalize the saliency score, we impose the constraints $0\leq\alpha, \beta\leq 1$ and $0\leq\alpha+\beta\leq 1$.

% The collection of salient relation triples can serve as the summary as well (Figure~\ref{fig:example}) since each relation description states a single piece of knowledge in a plausible way to some extent.
% Thus, we present \textbf{\proposedext}, an extractive summarization method that selects relation triples instead of sentences.

\subsection{Information Sentencification}
\label{subsec:steptwo}
For the generation of sentences from the selected relation triples (i.e., sentencification), \proposed builds a \textit{relation combiner} based on a pretrained text-to-text language model, such as \bart~\cite{lewis-etal-2020-bart} and  \tfive~\cite{raffel2020exploring}.
Using the relation combiner, \proposed can perform the abstractive summarization by sentencifying the selected relation triples.
%, and it is named \textbf{\proposedabs}.

\smallsection{Relation Combiner Training}
The relation combiner is effectively optimized with the self-supervised objective for the sentencification task.
To be specific, we collect training pairs of (relation triples, sentences) by randomly sampling a couple of sentences from source documents and extracting the relation triples from the sentences.
Then, we train the relation combiner based on Maximum Likelihood Estimation (MLE) to generate the sentences by taking the concatenated text of all the extracted relation triples.
As a result, it is expected to learn how to introduce linking words, place each component in order considering their relation, and remove duplicated phrases or entities, for plausible sentence generation. To eliminate redundancy, we apply a lightweight string similarity algorithm, Gestalt pattern matching \cite{ratcliff-etal-1988-pattern}, as a filter before merging relation triples.

\smallsection{Training Pair Filtering}
Despite the benefits of self-supervised training, the relation combiner still has a risk of introducing information that is not presented in a source document (i.e., extrinsic hallucinations) or factual errors against the document (i.e., intrinsic hallucinations) into its output summary.
Since the extracted relation triples are not guaranteed to perfectly cover all the information of their source sentences, some training pairs might guide the combiner to generate missing information that does not exist in the input relation triples.
To alleviate these hallucinations, we selectively collect the training pairs whose extracted relation triples fully cover the content of the source sentences.
Precisely, the pairs of (relation triples, sentences) are excluded from the training set, in case that some of the semantic tokens (i.e., nouns, proper nouns, and verbs) in the source sentences do not appear in the extracted relation triples.

\label{sec:framework}

\section{Demo: Interpretable Summarizing Tool}

\setlength{\fboxsep}{0pt}
\begin{figure*}[t]
    \centering
    \fbox{\includegraphics[width=\linewidth]{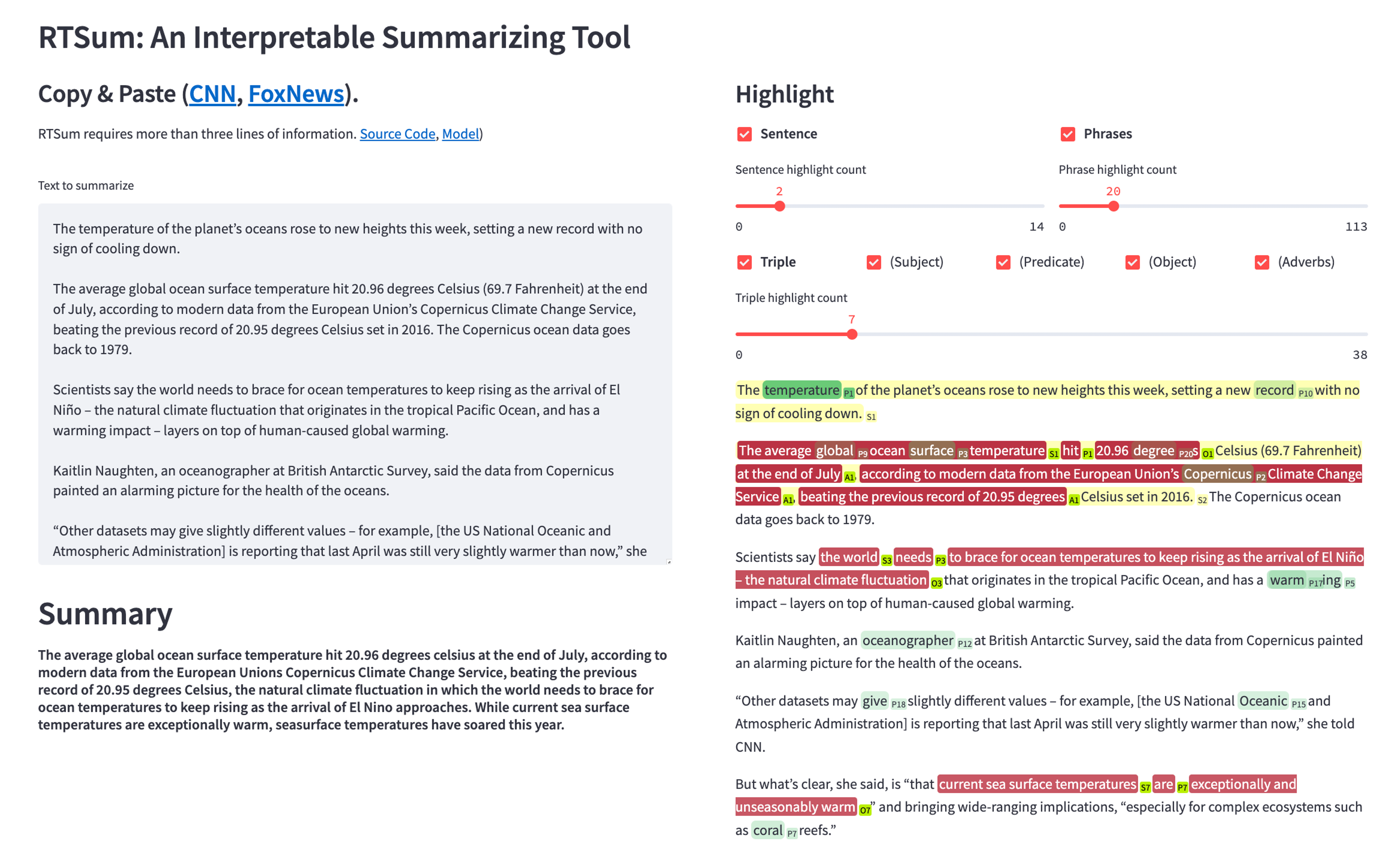}}
    \caption{Our interpretable summarizing tool features multi-level salience visualization. 
    Sentences, relation triples, and phrases with a high score are highlighted in yellow, red, and green, respectively. The saliency of each unit is denoted by its opacity.
    Within each triple, the subjects, predicates, objects, and adverbs are distinguished.}
    \label{fig:example}
\end{figure*}

Based upon our \proposed framework, we build an interpretable summarizing tool that provides not only the final summary of an input document but also its fine-grained interpretations.

\subsection{Multi-level Salience Visualization}
For interpretation of final summaries, our tool provides salience visualization for textual information units with different granularity (Figure~\ref{fig:example}).
It highlights the text spans that correspond to each information unit according to its salience score. 
For enhanced insights, the salience rank is explicitly annotated next to each span, providing users with a clear understanding of the relative importance of each information unit within the document. 
This feature allows users to grasp the significance of the textual content and gain a more nuanced and detailed understanding of the document's key points.

Users can personalize the salience visualization based on their unique preferences and specific needs, including customization options as follows:
\begin{itemize}
    \item \textbf{Type of textual units}: 
    Users have the flexibility to choose whether to highlight each type of textual unit. 
    They can opt to further dissect the highlight for a relation triple, differentiating its subject, predicate, and object components.
    \item \textbf{Number of textual units}: 
    Users can manually adjust the number of highlighted instances for each type of textual unit.
\end{itemize}

\subsection{Implementation Details}
\smallsection{Text graph construction}
For more reliable summarization, our \proposed implementation filters out less confident relation triples among the ones extracted from the OpenIE 5 system;
only the relation triples of which confidence is larger than 0.7 is considered as valid units.
To construct sentence-level and relation-level text graphs,
%(Sections~\ref{subsubsec:sensal} and \ref{subsubsec:relsal}), 
\proposed utilizes General Text Embeddings (GTE)\footnote{\texttt{https://huggingface.co/thenlper/gte-large}} as the sentence encoder, which is trained on a large-scale corpus of relevance text pairs covering a wide range of domains and scenarios.
Cosine similarity between two sentence (or relation description) embeddings is used for the edge weight in the graphs.

\smallsection{Relation triple selection}
\proposed in our tool simply ranks relation triples by their final salience scores, which are calculated by summing three distinct salience scores (i.e., $S^\text{s}, S^\text{r}, S^\text{p}$) with the same weight, and then chooses top-$K$ ones.
The number of relation triples to be selected is set to $K=3$. 

\smallsection{Relation combiner training}
To build a relation combiner, we fine-tune BART\footnote{\texttt{https://huggingface.co/facebook/bart-base}}~\cite{lewis-etal-2020-bart} to generate source sentences from the relation triples extracted from the sentences.
We use a text corpus in the news domain, {\cnndm}~\citep{nallapati-etal-2016-abstractive} which contains 287,113 news articles available for training.
To reduce the risk of hallucination, we filter out the cases that the amount of information in an input text (i.e., a set of relation triples) is shorter than that in an output text (i.e., sentences), as explained in Section~\ref{subsec:steptwo}.
In addition, we use three special tokens, \texttt{<subject>}, \texttt{<predicate>}, and \texttt{<object>}, to separate three components of each relation triple in an input text, which effectively provides structured information about each triple to the model.

\smallsection{Relation combiner alternatives}
While our summarizing tool provides the fine-tuned text-to-text language model as a default relation combiner, it also provides an option to employ instruction-following language models, such as InstructGPT~\cite{ouyang2022training} and ChatGPT.
These models can reconstruct plausible sentences from a set of relation triples, when being asked with a proper prompt written in natural language; 
they can be beneficial in that domain-specific or task-specific fine-tuning process is not required.

\label{sec:demo}

\section{Related Work}

\subsection{Unsupervised Extractive Summarization}
The most popular approach to unsupervised extractive summarization is to identify key sentences by using a text graph that represents the semantic (or lexical) relationship among text units in a source document.
% , whose node represents each sentence and edge models the semantic relevance among them~\cite{}.
TextRank~\cite{mihalcea-tarau-2004-textrank} is the first work to adopt a graph-based ranking algorithm~\cite{brin1998anatomy} to calculate the centrality of sentences in the graph, whose node represents each sentence and edge is modeled as the similarity between two sentences.
Several variants of TextRank have been implemented by utilizing symbolic sentence representations (e.g., TF-IDF)~\citep{barrios2016variations} or distributed sentence representations (e.g., skip-thoughts)~\cite{kiros2015skip} for computing the sentence similarity.

Most recent studies have employed pretrained language models (PLMs), such as BERT~\cite{devlin-etal-2019-bert}, to effectively model the salience of each sentence.
\citet{zheng-lapata-2019-sentence, lu2021unsupervised} used the degree-based node centrality of the position-augmented sentence graph where the sentence similarity is calculated by PLMs, and \citet{padmakumar2021unsupervised} defined the selection criterion by using PLM-based pointwise mutual information.
\citet{xu-etal-2020-unsupervised} considered the sentence-level self-attention score as the salience, after optimizing PLMs via masked sentence prediction.
Nevertheless, all of them regard a sentence as the basic unit for summarization, so they cannot exclude unnecessary information from each selected sentence. 

\subsection{Unsupervised Abstractive Sumamrization}
To train a neural model for abstractive summarization without using human-annotated text-summary pairs, most existing methods have adopted the auto-encoding architecture whose encoder compresses a source text into a readable summary (i.e., a few sentences) and decoder reconstructs the original text from the summary~\cite{wang-lee-2018-learning, baziotis-etal-2019-seq, chu2019meansum}.
Another line of research has focused on zero-shot abstractive summarization, which takes advantage of large-scale PLMs trained on massive text corpora.
Their models are optimized with a self-supervised objective (e.g., gap sentence generation)~\cite{raffel2020exploring, zhang2020pegasus} or heuristically-generated references (e.g., lead bias)~\cite{yang-etal-2020-ted, fang-etal-2022-leveraging}.
However, the well-known caveat of abstractive summarization is poor interpretability, which is also related to the hallucination problem;
their output summaries mostly contain factual errors or misinformation against the source document~\cite{kryscinski-etal-2020-evaluating, maynez-etal-2020-faithfulness}.

% \smallsection{Zero-shot or fine-tune on Lead Bias}
% lead bias without further tuning the language model for the summarization task (i.e., zero-shot summarization) or using lead bias obtained from large news corpora.

%% No publication yet
% \smallsection{Knowledge-Guided summarization}
% The most recent work on this line of research is decompose the overall summarization process into the two steps: extraction step and abstraction step.
% key phrases or events from a source document~\cite{}, and then such phrase-guided or event-guided summarization task based on Seq2Seq model, like \bart~\cite{}

% Meta-evaluation:
% \cite{chen-etal-2021-factuality-checkers}

% Triple-based:
% \citep{goodrich2019assessing}

% QA-based: % FEQA
% \citep{wang-etal-2020-asking, nan-etal-2021-improving}

% Pair classification-based:
% \citep{kryscinski-etal-2020-evaluating, zhang-etal-2021-fine-grained}

% Dependency arc classification-based:
% \citep{goyal-durrett-2020-evaluating, goyal-durrett-2021-annotating}
\label{sec:relatedwork}

\section{Conclusion}

In this paper, we introduce a summarization framework, called \proposed, which leverages relation triples as the basic units for summarization. 
Building upon this framework, we have developed a web demo for an interpretable summarizing tool that effectively visualizes the salience of textual units at three distinct levels.
Through our multi-level salience visualization, users can easily identify textual units impacting the summary and gain insights into the document's salient semantic structure.

Our \proposed framework and its user-friendly tool can effectively capture the essence of a document while maintaining interpretability.
The fusion of extractive and abstractive approaches, coupled with intuitive multi-level visualization, holds promise for applications requiring succinct, accurate, and interpretable summaries.
\label{sec:conclusion}

\section{Limitations}
Our study has the following limitations: 
Firstly, compared to single-step summarization approaches, our framework is relatively slower due to its multi-step process. 
Secondly, the current implementation relies on English-specific tools for sentence splitting and relation extraction, limiting its applicability to only English inputs. 
% However, the framework could potentially be extended to other languages by integrating appropriate information extraction tools for those languages. 
Lastly, while our research focuses on summarizing news articles effectively, the robustness and performance of our approach on longer or differently formatted text genres, such as books or research papers, has not been comprehensively evaluated.
\label{sec:limitations}

\section*{Acknowledgements}
This work was supported by the IITP grant funded by the Korea government (MSIT) (No.2020-0-01361) and the NRF grant funded by the Korea government (MSIT) (No. RS-2023-00244689).
\label{sec:ack}

\bibliography{bibliography}
\bibliographystyle{acl_natbib}

% \appendix

\end{document}